\documentclass[letterpaper, 10 pt, conference]{ieeeconf}  % Comment this line out if you need a4paper

\IEEEoverridecommandlockouts                              % This command is only needed if 
\usepackage{graphics} % for pdf, bitmapped graphics files
\usepackage{epsfig} % for postscript graphics files
\usepackage{amsmath} % assumes amsmath package installed
\usepackage{amssymb}  % assumes amsmath package installed
\usepackage{xcolor}
\usepackage{acronym}
\usepackage[scriptsize]{subfigure}

\newcommand{\ReTwo}{\mathbb{R}^2}
\newcommand{\Reals}{\mathbb{R}}

\newcommand{\Cspace}{Q}

\newcommand{\Cobs}{\Cspace_{obs}}

\newcommand{\Xspace}{X}
\newcommand{\Xfree}{\Xspace_{free}}
\newcommand{\Xobs}{\Xspace_{obs}}

\newcommand{\normdist}{f_\mathcal{N}}%{\mathcal{N}}
\newcommand{\RRTX}{$\text{RRT}^\text{X}$}
\newcommand{\RRTst}{$\text{RRT}^*$}
\newcommand{\G}{\mathcal{G}}
\newcommand{\T}{\mathcal{T}}
\newcommand{\GMM}{\scriptscriptstyle GMM}
\newcommand{\GMR}{\scriptscriptstyle GMR}

\acrodef{2D}[2D]{2-dimensional}
\acrodef{GMR}[GMR]{Gaussian Mixture Regression}
\acrodef{GMM}[GMM]{Gaussian Mixture Model}
\acrodef{EM}[EM]{Expectation Maximization}
\acrodef{pdf}[pdf]{probability density function}

\DeclareMathOperator{\lmc}{lmc}
\DeclareMathOperator{\cost}{cost}
\DeclareMathOperator{\DI}{DI}
\DeclareMathOperator{\PDI}{PDI}
\DeclareMathOperator{\proj}{proj}
\DeclareMathOperator*{\argmin}{arg\,min}

\title{\LARGE \bf
Safe motion planning for a mobile robot navigating\\in environments shared with humans
}

\author{Basak Sakcak$^{1}$ and Luca Bascetta$^{2}$% <-this % stops a space
\thanks{$^{1}$Center of Ubiquitous Computing, Faculty of Information Technology and Electrical Engineering, University of Oulu, Oulu, Finland, {\tt\small basak.sakcak@oulu.fi}}
\thanks{$^{2}$Politecnico di Milano, Dipartimento di Elettronica, Informazione e Bioingegneria, Piazza L. Da Vinci 32, 20133, Milano, Italy, {\tt\small luca.bascetta@polimi.it}}%
}

\begin{document}
\maketitle
\thispagestyle{empty}
\pagestyle{empty}

%%%%%%%%%%%%%%%%%%%%%%%%%%%%%%%%%%%%%%%%%%%%%%%%%%%%%%%%%%%%%%%%%%%%%%%%%%%%%%%%
\begin{abstract}
In this paper, a robot navigating an environment shared with humans is considered, and a cost function that can be exploited in \RRTX, a randomized sampling-based replanning algorithm that guarantees asymptotic optimality, to allow for a safe motion is proposed. The cost function is a path length weighted by a danger index based on a prediction of human motion performed using either a linear stochastic model, assuming constant longitudinal velocity and varying lateral velocity, and a GMM/GMR-based model, computed on an experimental dataset of human trajectories. The proposed approach is validated using a dataset of human trajectories collected in a real world setting.
\end{abstract}
%%%%%%%%%%%%%%%%%%%%%%%%%%%%%%%%%%%%%%%%%%%%%%%%%%%%%%%%%%%%%%%%%%%%%%%%%%%%%%%%
\section{Introduction}
Autonomous mobile robots have been increasingly sharing environments with humans in many different applications, ranging from service and goods delivery, to care-giving. This trend brings in the additional burden of being accepted by humans, forcing robot navigation system to guarantee human safety and comfort, as well. Safety requires that collisions with humans or other obstacles are avoided, while comfort forces robots to limit velocity close-by humans and to keep an appropriate distance respecting human private space.%\\
In conventional motion-planning problems, obstacles are usually assumed as static, i.e., they do not move while robot follows the planned trajectory, or they move slowly with respect to the robot, or their motion can be measured and predicted. Humans, instead, are expected to move at a velocity that is comparable with robot speed, and their motion, though measurable, is highly unpredictable. Therefore, a motion-planning algorithm for such dynamic environments should be able to efficiently \emph{replan}, so as to ensure that the computed trajectory is collision-free for the entire robot motion. A viable solution, allowing to cope with human motion unpredictability and replanning complexity, can be devised using \RRTX~\cite{otte2016RRTX}, allowing to repair graph unfeasible parts, and introducing a prediction of human motion that, under mild assumptions, holds for short time windows.

In this work, the problem of motion-planning for an autonomous mobile robot in environments shared with humans, focusing on human-avoidance, is addressed. \RRTX\; is adopted as the core motion-planning algorithm, and a human-aware cost function that encodes the path-length (a standard optimization metric for motion-planning) and the danger of colliding with a human is introduced. Assuming that human position and orientation are known to the robot at each time instance, two models to predict human motion within a short period of time, hence to assess the potential danger, are introduced. Whereas the first one is a linear stochastic model that derives from~\cite{de2007onboard}, the second model%, instead, 
follows a learning-by-demonstration approach~\cite{calinon2007learning}.%\\
The proposed approach is validated on a dataset of pedestrians moving along a sidewalk. 

Most replanning algorithms for mobile robots, like D*~\cite{stentz1995focussed}, LPA*~\cite{koenig2004lifelong}, and D*-Lite~\cite{koenig2005fast}, incrementally build a graph whose nodes lie on a 2D grid, while repairing parts of it affected by potential collisions with unknown or dynamic obstacles through heuristics. Sampling-based algorithms, that randomly sample the continuous robot configuration space, are used for replanning too. These algorithms (consider e.g.,~\cite{ferguson2006replanning,zucker2007multipartite}) are based on RRT and repair tree affected parts similarly to D*. \RRTX~\cite{otte2016RRTX} was the first asymptotically optimal replanning algorithm, that combines graph rewiring property of \RRTst~for optimality and repairing property of D* for replanning. Other works based on \RRTst~rely on heuristic-based sampling to compensate for nodes that fall in obstacles, and rewire the tree locally~\cite{adiyatov2017novel,connell2017dynamic}.    

When the environment is known with uncertainty, a standard approach is to constrain the path to have a pre-defined clearance~\cite{geraerts2010planning,wang2020neural}, or to ensure that it is as far from obstacles as possible~\cite{denny2014marrt}. An exact method to compute all paths that minimize length and maximize clearance is reported in~\cite{sakcak2021complete}. 
For disc obstacles with bounded speed,~\cite{van2008planning} proposes an exact algorithm to compute the shortest path through cones. Finally, considering probabilistic uncertainties, chance-constrained formulations were used~\cite{luders2010chance}. Despite being powerful, these methods cannot generalize to scenarios where new obstacles appears during robot motion, which is expected in the case considered in this paper. 

All of the previously mentioned replanning algorithms can be used in environments shared with humans.
Without any knowledge of potential changes in the environment, nodes and edges can be added to graph corresponding to robot configurations that are currently collision-free but become invalid once the environment changes. This issue causes the algorithm to continuously replan, which is a computationally intensive procedure. 
On the other hand, human motion is not completely unpredictable (interested readers can refer to~\cite{rudenko2020human} for a recent survey on human motion prediction), and short-term motion can be estimated using simple techniques \cite{rudenko2020human}. Therefore, this work combines a replanning algorithm with human motion prediction deriving from previous works \cite{de2007onboard,calinon2007learning} to introduce a danger index penalizing nodes and edges that can potentially result in collisions, not only considering the present but also the future, reducing the need for replanning operations.
A sampling-based algorithm, that can tackle higher dimensions with relative ease, is selected to search the robot configuration space augmented with a time-dimension, so that only the parts of the graph falling into areas that can be occupied by a person at compatible times are penalized.

\section{Problem Formulation}
This work concerns a holonomic mobile robot moving in a 2D planar environment, whose configuration is represented by $q=(q_x, q_y) \in \Cspace \subset \ReTwo$, $\Cspace$, assumed to be a compact set, being the robot configuration space. Let $T \subset \Reals$ be an interval of time over which the problem is defined. Without loss of generality, the robot starts its motion at $t=0 \in T$. Assuming the problem does not impose any restrictions on time to reach the goal, $T=[0,\infty)$ can be selected. 

The environment is characterised by static and moving obstacles, i.e., humans, that are represented as open subsets of $\ReTwo$. Obstacles belonging to different classes are assumed to be distinguishable to the robot, and the robot has partial knowledge of the environment, i.e., positions of certain static obstacles are known a priory, whereas other static obstacles and humans are discovered through robot sensors as the robot moves. 
Therefore, the obstacle space (the union of all obstacles) $\Cobs(t)$ can be defined as the set of configurations that are forbidden to the robot at time $t \in T$ due to collisions. Since $\Cobs$ varies with time, the robot state space $X$ is defined as the Cartesian product $X= \Cspace \times T$, and the set of states $\Xobs$ that are not reachable by the robot due to collisions are defined as $\Xobs(t) = \{(q,t) \in \Xspace \mid q \in \Cobs(t)\}$.
Consequently, the set of states that the robot is not in collision with obstacles is denoted by $\Xfree=\Xspace\setminus\Xobs$.\\
Furthermore, $\Delta\Xobs$ is used to denote a change in the environment as a function of time and robot state. 

A trajectory connecting $x_0$ to $x_1$ is described as $\pi : [0,1] \rightarrow X$ such that $0 \mapsto x_0$ and $1 \mapsto x_1$, and denoted by $\pi(x_0,x_1)$. $\pi(x_0,x_1)$ is collision-free iff $\pi(x_0,x_1) \cap \Xobs = \emptyset$. The length of $\pi(x_0,x_1)$ is denoted by $d_\pi(x_0,x_1)$.%\\
A cost function $\cost$ attributes a cost to each trajectory $\pi(x_0,x_1)$, denoted by $\cost_\pi(x_0,x_1)$.

Let $x_{start}$ be the start state of the robot corresponding to $t=0$ and $q_{goal} \times T$, in which $q_{goal}$ is the goal position, be a goal region (considering no bounds on the arrival time). Let $\Pi(x_{curr})$ be the set of all collision-free paths starting at $x_{curr}$, the current robot state, and ending in the goal region.\\
Given $X,\Xobs(0), x_{goal}$, and  $x_{start}$, with $\Delta\Xobs$ unknown, the minimum-cost replanning problem is defined as simultaneously updating the robot state along $\pi^*(x_{curr},x_{goal})$ while recalculating $\pi^*(x_{curr},x_{goal})$ when $\Delta\Xobs \neq \emptyset$, in which
\begin{displaymath}
    \pi^*(x_{curr},x_{goal}) = \argmin_{\pi \in \Pi(t)} \cost_\pi(x_{curr},x_{goal})
\end{displaymath}
is the optimal trajectory.

\section{Human Motion Models}\label{sec:hum_models}
This section introduces the two motion models that are used to predict the human motion.

\subsection{Model based on~\cite{de2007onboard}}\label{sec:hum_model1}
The first model is a linear stochastic model (SM for short) that is based on~\cite{de2007onboard}. % and it is defined as follows
%The first model here considered is based on~\cite{de2007onboard}, which assumes linear dynamics.
It is defined as follows
\begin{equation}\label{eq:linear_model}
    \begin{aligned}  
      p_{ln}(k+1) &= p_{ln}(k) + \delta t\, v_{ln}(k) \\
      v_{ln}(k+1) &= v_{ln}(k)\\
      p_{lt}(k+1) &= p_{lt}(k) + \delta t\, v_{lt}(k) \\
      v_{lt}(k+1) &= v_{lt}(k) + \delta t\, w_{lt}(k),
    \end{aligned}
\end{equation}
in which $p_{ln}$, $p_{lt}$ and  $v_{ln}$, $v_{lt}$ are longitudinal and lateral positions and velocities, $\delta t= 0.1\,\mathrm{s}$, and $w_{lt}$ is a white Gaussian noise with zero mean and variance $\sigma^2_{lt}=0.5$. Differently from~\cite{de2007onboard}, a constant deterministic velocity in the longitudinal direction, $v_{ln}(k)=1.2\,\mathrm{m/s}$ for all $k$, is assumed. The lateral dynamics can be represented as 
\begin{equation*}
    x_{lt}(k+1)=A x_{lt}(k) + G w_{lt}(k)
\end{equation*}
in which $x_{lt}=[p_{lt}, v_{lt}]^T$, and
\begin{equation*}
    \begin{aligned}
        A=\begin{bmatrix}
            0 & \delta t \\
            0 & 1
        \end{bmatrix} & &
        G=\begin{bmatrix}
        0 \\ \delta t
        \end{bmatrix}.
    \end{aligned}
\end{equation*}
The covariance matrix corresponding to $Gw_{lt}(k)$ is 
\begin{equation*}
    \Sigma_w=\begin{bmatrix}
    0 & 0 \\
    0 & \delta t^2\sigma_{lt}^2
    \end{bmatrix}.
\end{equation*}
Suppose that $x_{lt}(0)=[0,0]^T$, and that there is no uncertainty in the initial lateral position and velocity, i.e., $\Sigma_{lt}(0)=0_{4\times4}$. Then, the covariance matrix can be iteratively computed as \begin{equation*}
    \Sigma_{lt}(k+1)=A \Sigma_{lt}(k) A^T + \Sigma_w.
\end{equation*}
The covariance matrix at stage $k$ is
\begin{displaymath}
    \Sigma_{lt}(k)=
    \begin{bmatrix}
    \sigma^2_p(k) & \star \\
    \star &  \star
    \end{bmatrix},
\end{displaymath}
in which $\sigma^2_p(k)$ is the variance of the lateral position at stage $k$. Note that the mean is always zero. %\\

Since a constant longitudinal speed is adopted, $\sigma^2_p(k)$ can be rewritten as a function of $p_{ln}$, such that $\sigma^2_p(p_{ln}(k))=\sigma^2_p(k)$, and we have a distribution of lateral positions for each $p_{ln}$. %\\
%The \ac{pdf} corresponding to $p_{lt}$ for each longitudinal position $p_{ln}$ is written as
Due to the normality assumption on the noise, the \ac{pdf} corresponding to $p_{lt}$ for each longitudinal position $p_{ln}$ is
\begin{equation}\label{eq:pdf_lat_gauss}
     f_{SM}(p_{lt} \mid p_{ln}) = \normdist(p_{lt}; 0, \sigma^2_p(p_{ln})),
\end{equation}
in which $\normdist(\cdot;\mu,\sigma^2)$ denotes the \ac{pdf} of a normal distribution with mean $\mu$ and variance $\sigma^2$.

\begin{figure}[htbp]
    \centering
    \includegraphics[width=0.6\linewidth]{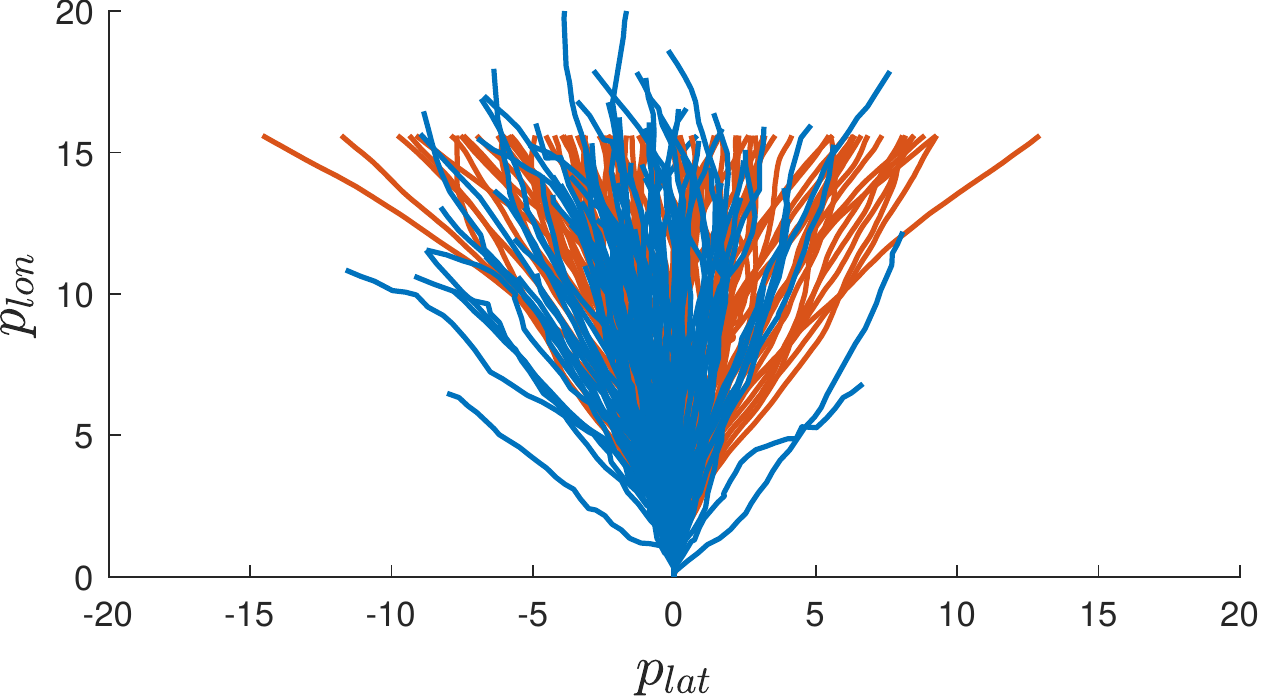}
    \vspace{-0.75em}
    \caption{Orange paths are generated by~\eqref{eq:linear_model}, blue paths are taken from the {BIWI} dataset~\cite{pellegrini2009you} corresponding to exiting a building.}
    \label{fig:rollout}
\end{figure}

Fig.~\ref{fig:rollout} shows 80 paths simulated for $13\,\textrm{s}$ using the model introduced in~\eqref{eq:linear_model} together with paths executed by pedestrians exiting a building, taken from~\cite{pellegrini2009you}. 

\subsection{Models based on GMM and GMR}
Since only the spatial properties of human trajectories are of interest, an observation corresponds to longitudinal and lateral coordinates. A dataset of $N$ observations is represented as $\{\xi_j\}_1^N$ in which $\xi_j=(p_{ln},p_{lt}) \in \Reals^2$ 
%is an observation corresponding 
corresponds to longitudinal and the lateral coordinates.   

A \ac{GMM} \ac{pdf} constituted by $K$ components is defined as $f_{\GMM}(\xi_{j}) = \sum_{k=1}^{K} \gamma_{k}f_{\GMM}(\xi_{j}\mid k)$ in which $\gamma_{k}$ is the prior and $f_{\GMM}(\xi_{j}\mid k)$ is the conditional \ac{pdf} satisfying $f_{\GMM}(\xi_{j}\mid k) = \normdist(\xi_j ; \mu_k, \Sigma_k)$. The parameters $\{\gamma_k, \mu_k, \Sigma_k\}$ of the (bivariate) Gaussian component $k$ correspond to the prior, mean, and covariance, respectively. The priors $\gamma_{k} \in [0,1]$ satisfy $\sum_{k=1}^{K} \gamma_{k} = 1$. %\\
Given a set of observations and the total number of components $K$, the parameters of a GMM, that are $\{\gamma_k, \mu_k, \Sigma_k\}_1^K$, are learned using the standard \ac{EM} algorithm. \ac{EM} requires an initial guess to be provided, which is derived from the clusters found by a $k$-means clustering technique, as recommended in~\cite{calinon2007learning}.

\ac{GMR} is used to reconstruct a general form for the data. Longitudinal coordinates are used as query points and corresponding lateral coordinates are estimated through regression. For each GMM, the parameters corresponding to the lateral and longitudinal coordinates are
\begin{equation*}
\mu_k=(\mu_{ln,k}, \mu_{lt,k})\qquad
\Sigma_k=
\begin{bmatrix}
\sigma^2_{ln,k} & \sigma_{lnlt,k} \\
\sigma_{lnlt,k} & \sigma^2_{lt,k} 
\end{bmatrix}.
\end{equation*}
Given $p_{ln}$, for each component $k$, the conditional expectation of $\widehat{p}_{lt,k}$ and the estimated conditional variance are $\widehat{p}_{lt,k} = \mu_{lt,k} + \tfrac{\sigma_{lnlt,k}}{\sigma^2_{ln,k}}(p_{ln}-\mu_{ln})$ and $\hat{\sigma}^2_{lt,k} =\sigma^2_{lt,k}-\tfrac{\sigma^2_{lnlt,l}}{\sigma^2_{lt}}$.
To consider all the GMM components, $K$ mixing coefficients, interpreted as the relative responsibility of each component, are defined as a function of the longitudinal position. Each mixing coefficient $\beta_{k}$, $k=1,\dots,K$ corresponds to
\begin{equation*}
\beta_{k} = \frac{\gamma_k\normdist(p_{ln};\mu_{ln,k}, \sigma^2_{ln,k})}{\sum_{i=1}^{K} \gamma_{ln,i} \normdist(p_{ln}; \mu_{ln,i}, \sigma^2_{ln,i})}.
\end{equation*}
Using mixing coefficients, given $p_{ln}$, the conditional expectation and the conditional variance can be computed as 
\begin{equation*}
\widehat{p}_{lt}=\sum_{k=1}^{K} \beta_{k}\hat{p}_{lt,{k}}\qquad
\hat{\sigma}^2_{lt}=\sum_{k=1}^{K} \beta_{k}^{2}\hat{\sigma}^2_{lt,{k}}.
\end{equation*}
By computing $\{\widehat{p}_{lt}, \hat{\sigma}^2_{lt}\}$ at different longitudinal coordinates the expected lateral coordinate and the associated variance can be obtained. The complete conditional distribution is 
\begin{equation}\label{eq:pdf_GMR}
    f_{\GMR}(p_{lt} \mid p_{ln})=\sum_{k=1}^K \beta_k \normdist(p_{lt}; \widehat{p}_{lt,k}, \hat{\sigma}^2_{lt,k}).
\end{equation}
\begin{figure}[htbp]
    \centering
    \subfigure[]{\includegraphics[trim=30 30 30 30,clip,width=.45\linewidth]{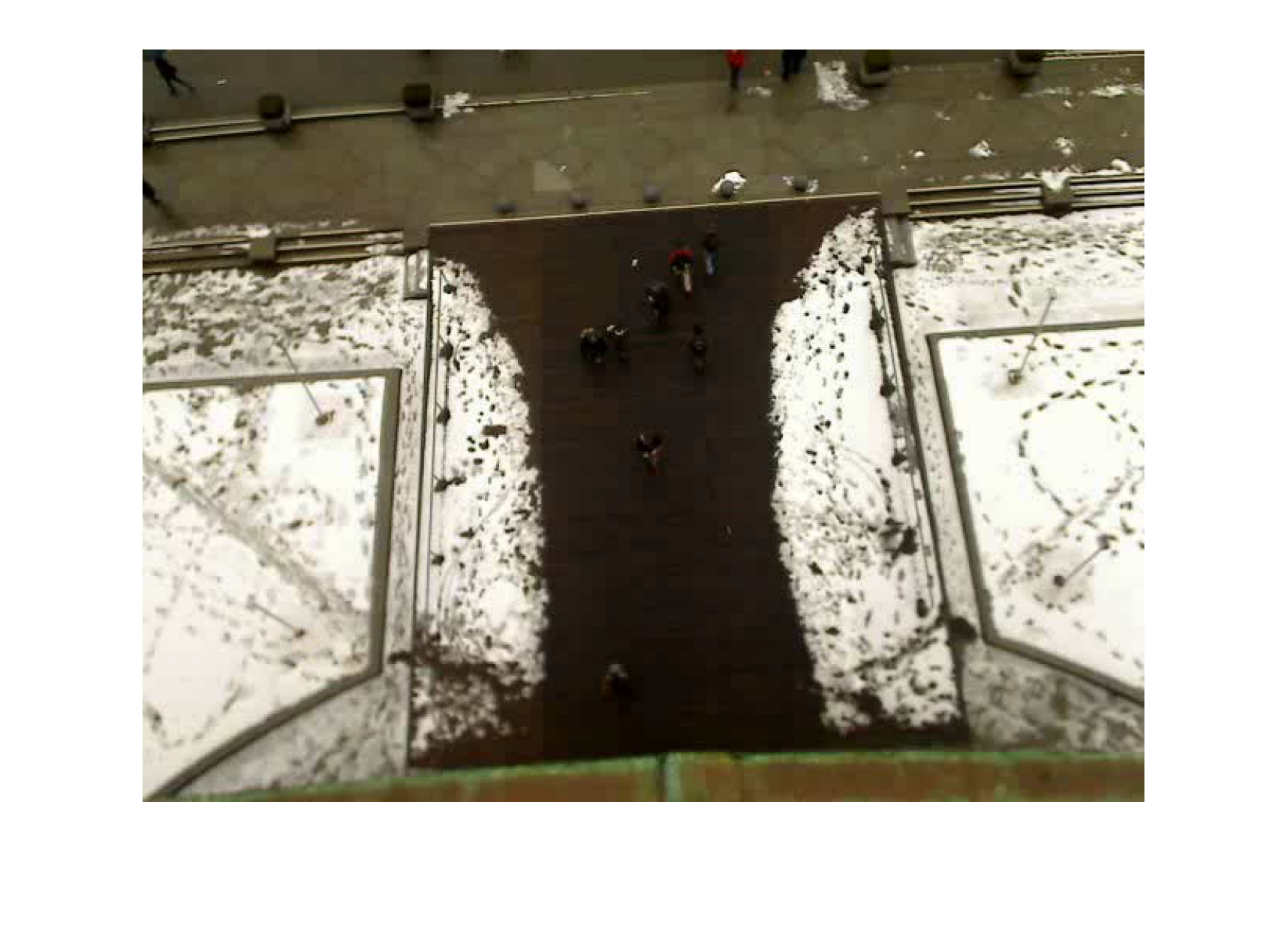}}
    \subfigure[]{\includegraphics[trim=20 20 30 30,clip,width=.45\linewidth]{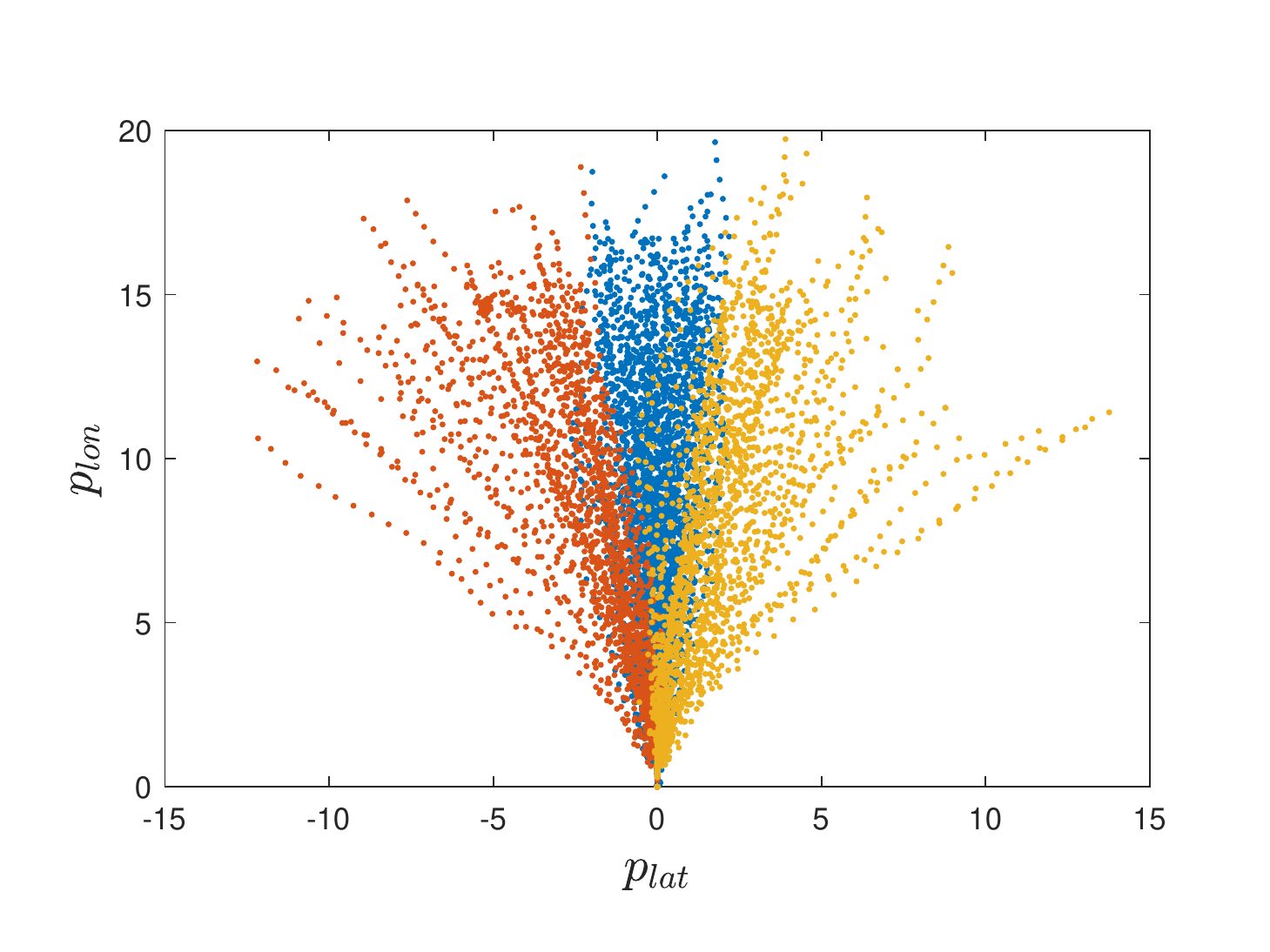}}
    \vspace{-0.75em}
    \caption{(a) Environment corresponding to the trajectories in the database. (b) Paths divided into three classes: straight (blue), rightward (yellow), leftward (orange). Units in meters.}
    \label{fig:three_classes}
\end{figure}
\begin{figure*}[htbp]
\centering
\subfigure[Straight]{\includegraphics[trim=10 100 10 100,clip,width=.25\linewidth]{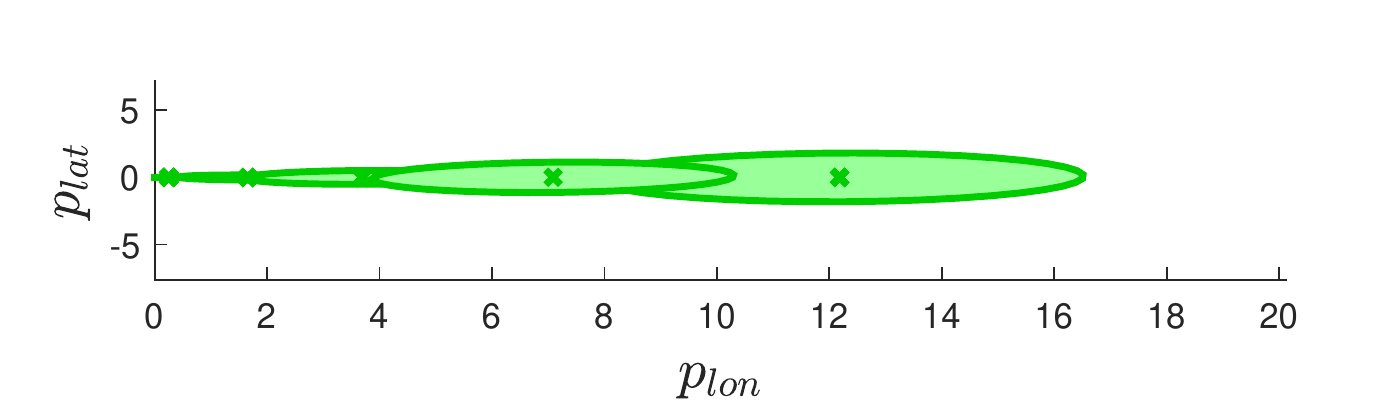}}
\subfigure[Rightward]{\includegraphics[trim=10 55 10 40,clip,width=.25\linewidth]{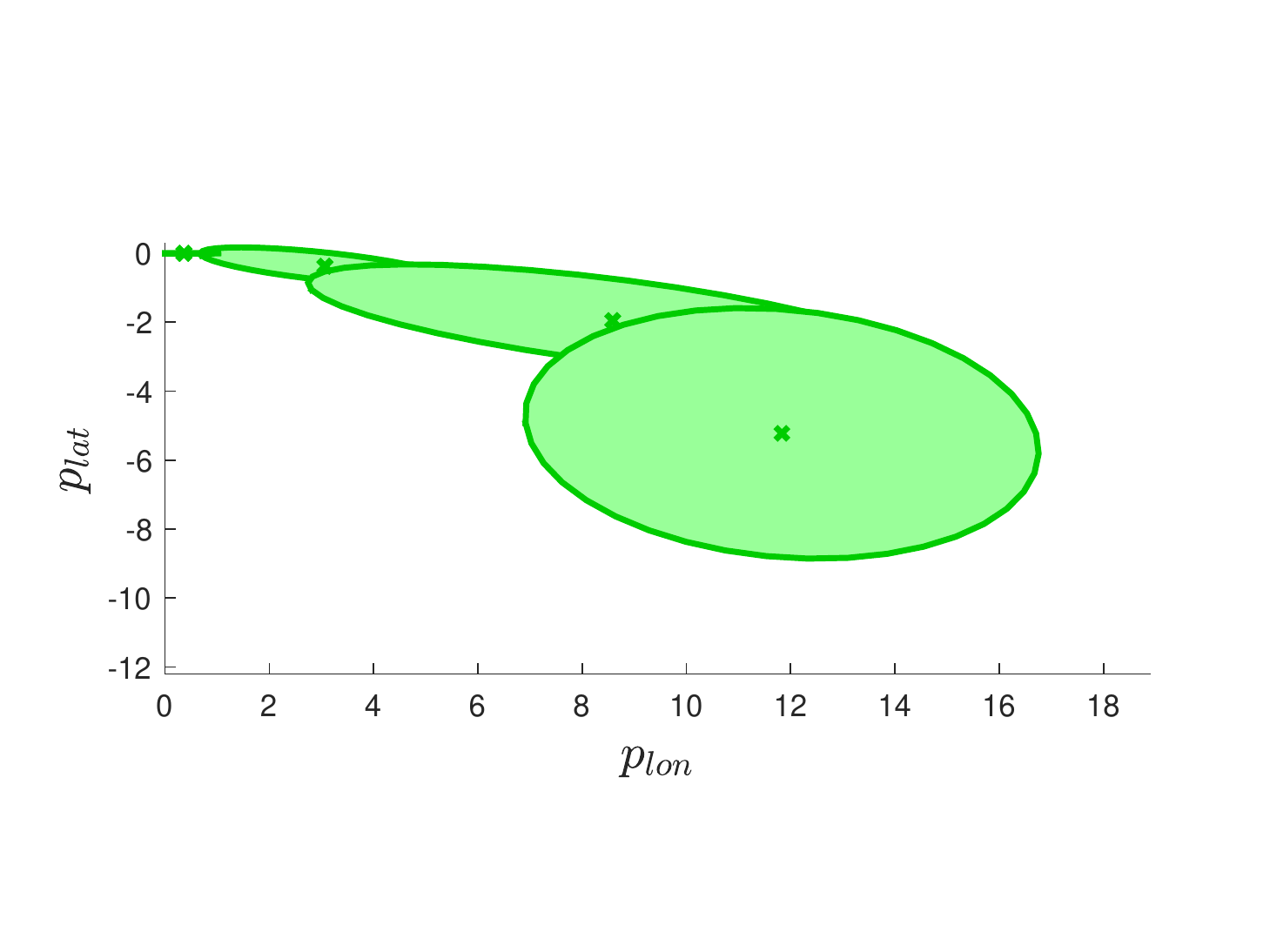}}
\subfigure[Leftward]{\includegraphics[trim=10 40 10 40,clip,width=.25\linewidth]{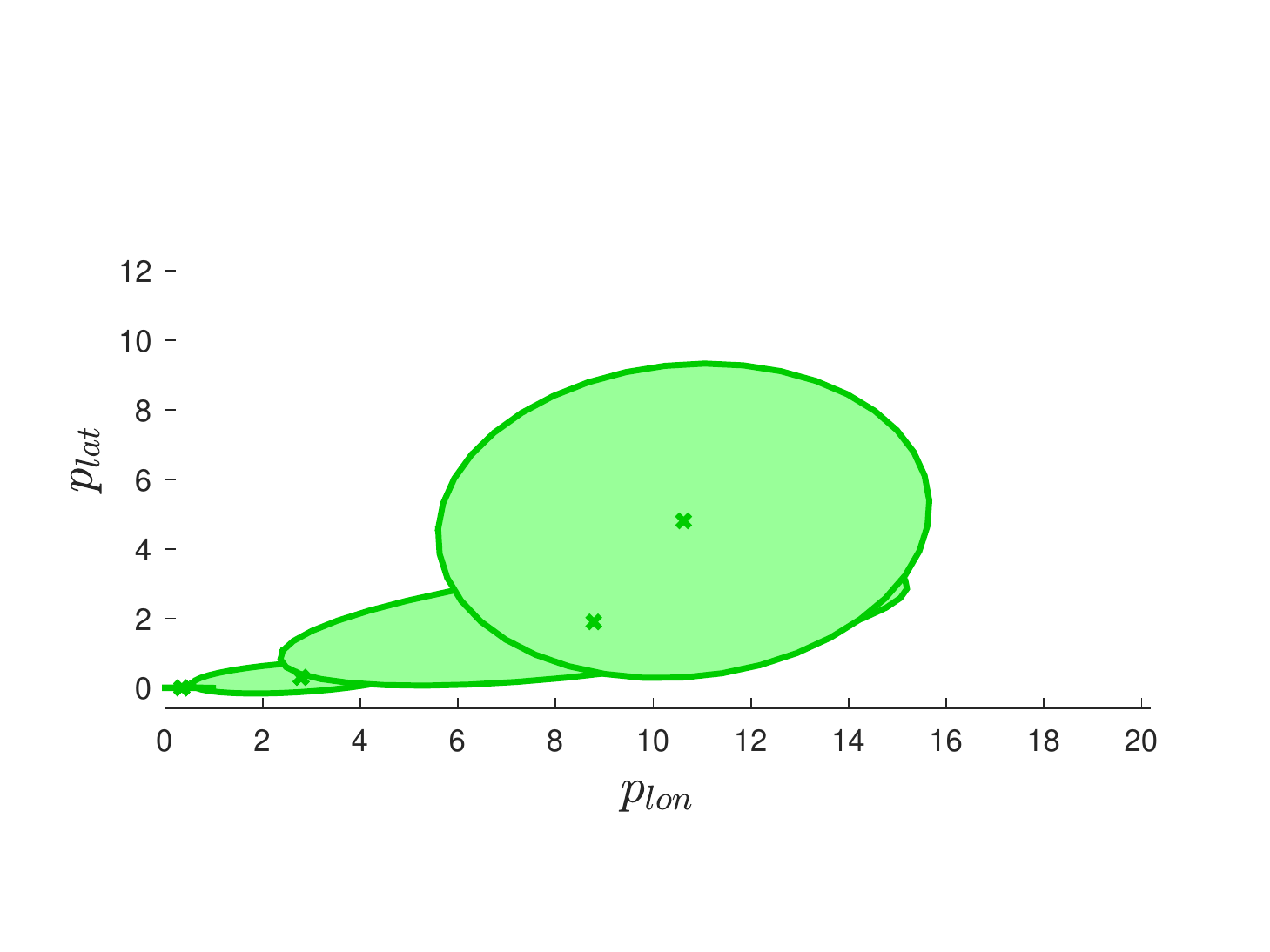}}\\
\subfigure[Straight]{\includegraphics[trim=10 100 10 100,clip,width=.25\linewidth]{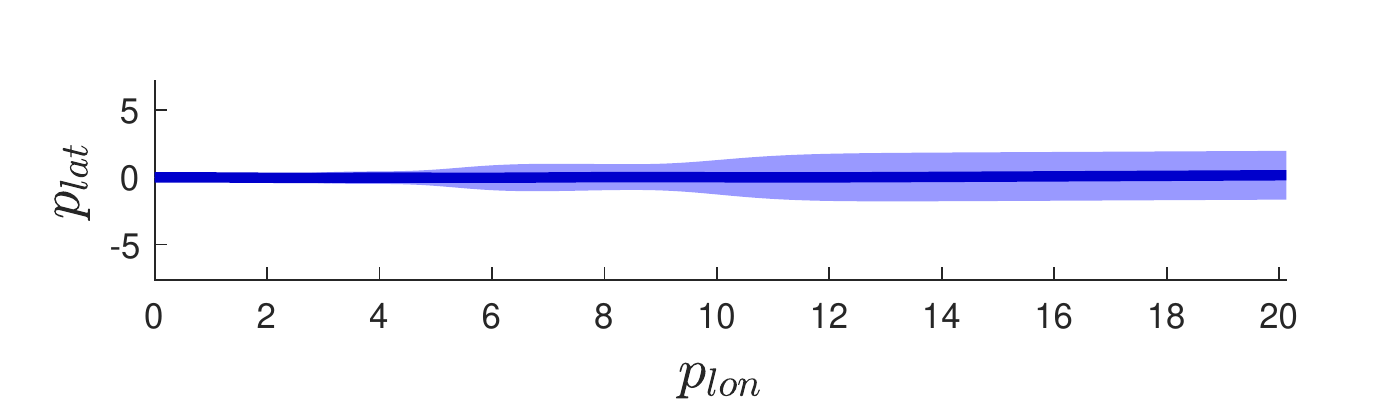}}
\subfigure[Rightward]{\includegraphics[trim=10 55 10 40,clip,width=.25\linewidth]{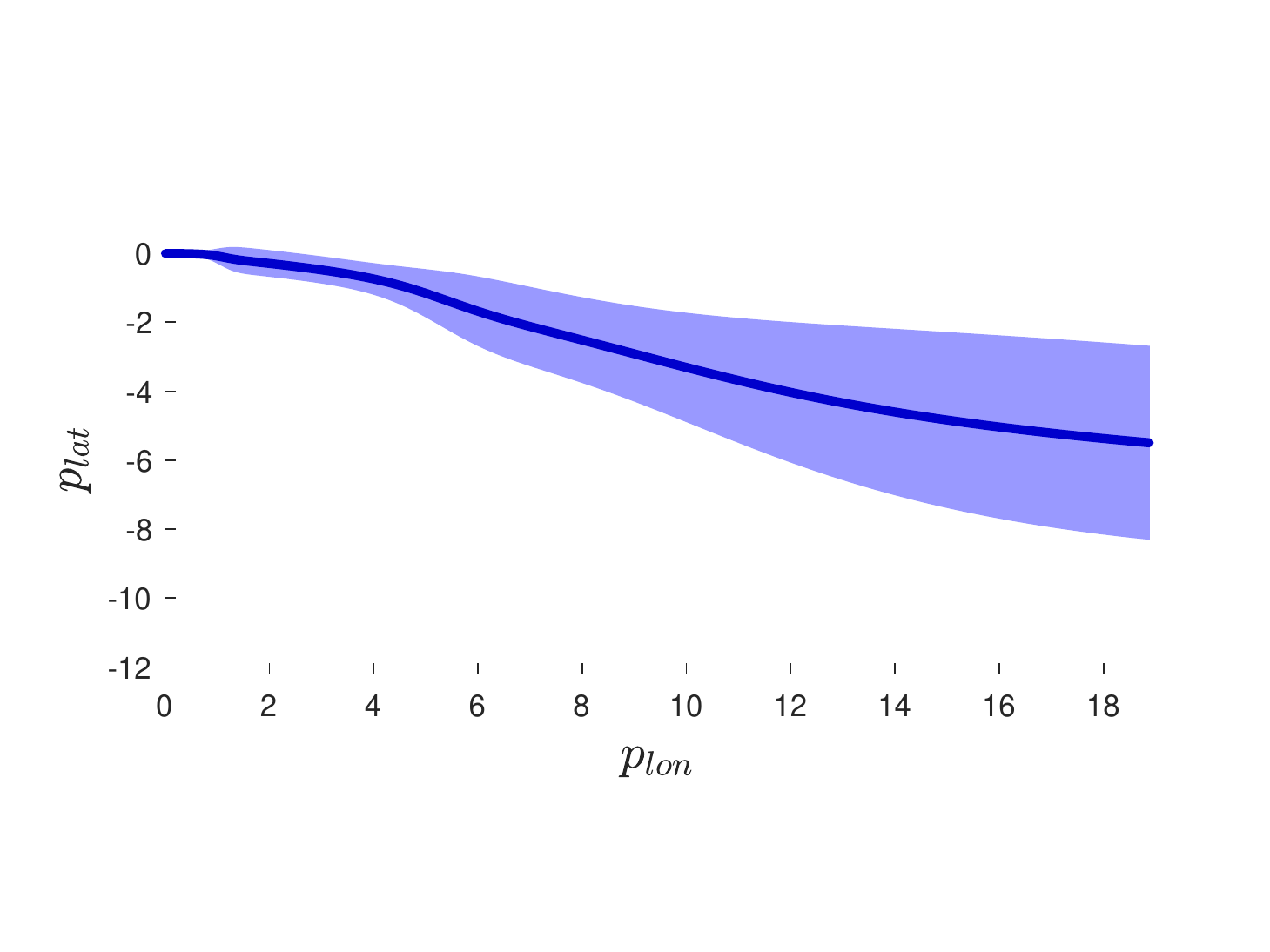}}
\subfigure[Leftward]{\includegraphics[trim=10 40 10 40,clip,width=.25\linewidth]{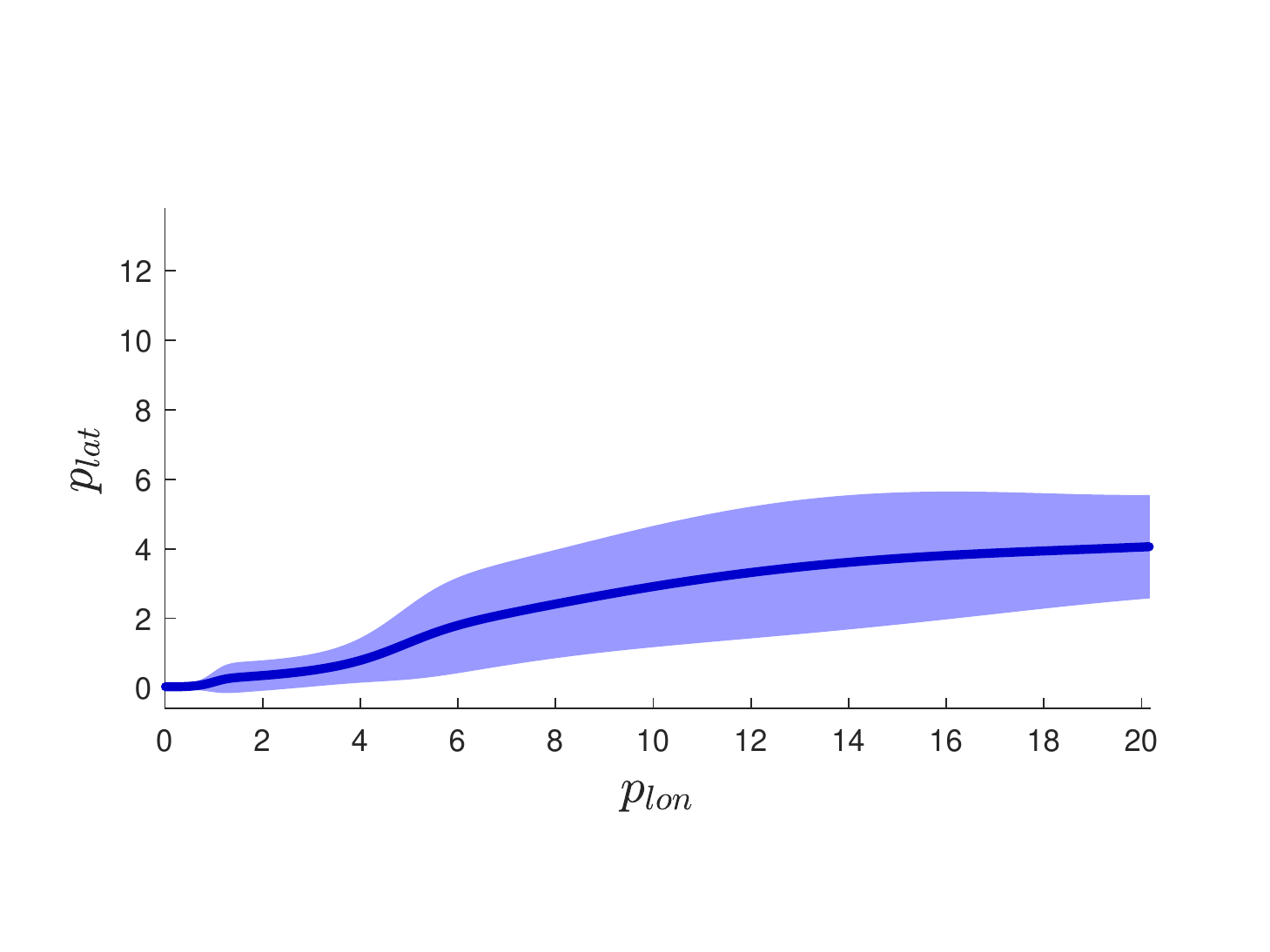}}
\vspace{-0.75em}
\caption{Models based on three classes of paths (straight, rightward, leftward). (a-c) Components of the GMMs, up to 3 standard deviations for each dimension. (d-f) Corresponding reference trajectories (purple line) and the lateral position variance, up to 3 standard deviations, computed using GMR.}
\label{fig:GMM_GMR}
\end{figure*}

To determine the parameters of the GMM/GMR model, the BIWI Walking Pedestrians dataset \cite{pellegrini2009you} has been used. In particular, paths executed by pedestrians near a university building are used to generate the GMM (see Fig.~\ref{fig:three_classes}a). %This way no strict restrictions are implied on the possible goal positions. 
Among all trajectories, only the ones compatible in duration have been considered, and the time information has been removed. Since a human-fixed reference frame is considered, the respective paths were all translated and rotated such that they start from the origin with initial orientation aligned with the positive direction along the longitudinal axis. %\\
Furthermore, analysing the dataset one can notice that the paths follow three main trends: straight, rightward, and leftward. Thus, in order to make more coherent human motion models, based on GMM and GMR methods, the dataset was divided in three parts (see Fig.~\ref{fig:three_classes}b), and the paths that do not follow one of the three main trends have been removed, i.e., the trajectories which take negative position values along the longitudinal axis. %\\
Finally, Fig.~\ref{fig:GMM_GMR} shows the normal distributions composing GMMs computed for each of the path classes, and the respective general path computed using GMRs.

To generate a single model that captures all three trends, all components have been combined in a single GMM in which the relative weights (priors) are scaled based on the normalized number of trajectories used to generate that model. Let $S$, $R$, $L$ be the labels for the GMMs corresponding to the straight, rightward, and leftward motions, respectively.
Let $\mathcal{F}=\mathcal{F}_S \cup \mathcal{F}_R \cup \mathcal{F}_L$ be the set of all components used for all three models such that each element has a unique label $i$, $i=1,\dots, K_T$ and $K_T=|\mathcal{F}_S|+|\mathcal{F}_R|+|\mathcal{F}_L|$ is the total number of components used for all three models. Let $\ell(i) \in \{S,R,L\}$ be the class label of the $i^{th}$ element of $\mathcal{F}$, and $\eta(i)\in[0,1]$ be the prior of the component in the respective GMM corresponding to one of the three models. The GMM corresponding to all three models is expressed as 
\begin{equation}\label{eq:GMM_all_three}
    f_{\GMM}(\xi_j)=\sum_{i=1}^{K_T}\gamma_i f_i
\end{equation}
in which $f_i \in \mathcal{F}_{\ell(i)}$ is the corresponding conditional \ac{pdf}. The respective normalized prior is computed as $\gamma_i=\eta(i)n_{\ell(i)}/n$, in which $n_{\ell(i)}$ is the number of paths used for generating the GMM corresponding to class $\ell(i)$, and $n=n_S+n_R+n_L$ is the total number of considered paths. In our case, all three sets have approximately the same cardinality meaning that there is no clear tendency towards one model. This is indeed expected. One could also dynamically change the relative contribution of each model, equivalent to estimating the motion trend. We leave this as a future work. 
Moreover, considering all the GMR parameters together, the relative likelihood of a position where the pedestrian could be in the future can be defined as follows
\begin{equation}\label{eq:pdf_GMR_all_three}
f_{\GMR}(p_{lt} \mid p_{ln}) = \sum_{i=1}^{K_T} \gamma_i f_{\GMR,\ell(i)}(p_{lt}\mid p_{long}), 
\end{equation}
in which $f_{\GMR,\ell(i)}$ is the conditional distribution computed as in \eqref{eq:pdf_GMR}, considering trajectories that belong to class $\ell(i)$.

To estimate the likelihood of colliding with a person, two models have been considered, one based on a GMM/GMR model, named \textit{Bivariate GMM}, and one on a \textit{Univariate GMR} model. 
The first model is based on~\eqref{eq:GMM_all_three}, capturing all three trends and assuming there is uncertainty affecting also the longitudinal position, the second model derives from~\eqref{eq:pdf_GMR_all_three} and considers the variance only in the lateral position.  

\subsection{Danger Index}\label{sec:danger_index}
Using the previously introduced human motion models, a danger index can be assigned to a robot position, capturing the likelihood of colliding with a human. Let $q=(q_x, q_y)$ be the robot position with respect to the absolute reference frame, and let $(q_{long},q_{lt})$ be the robot position expressed in the coordinate frame of the human. %\\

The danger index $\DI(q_{ln},q_{lt})$ of a position $(q_{ln},q_{lt})$ corresponds to the probability of being in the $\varepsilon$-vicinity of a human, in which $\varepsilon$ is a small number, and is computed as 
\begin{equation} \label{eq:DI_lin_sto_model}
\DI_{SM}(q_{long},q_{lt})=\int_{q_{lt}-\varepsilon}^{q_{lt}+\varepsilon}f_{SM}(p_{lt} \mid q_{ln}) dp_{lt}
\end{equation}
for the linear stochastic motion model, in which $f_{SM}$ is defined in~\eqref{eq:pdf_lat_gauss}, or similarly as
\begin{displaymath}
\DI_{\GMM}(q_{ln},q_{lt})=\int_{q_{lt}-\varepsilon}^{q_{lt}+\varepsilon}\int_{q_{ln}-\varepsilon}^{p_{ln}+\varepsilon}f_{\GMM}(p_{ln},p_{lt})dp_{lt}dp_{ln}
\end{displaymath}
using~\eqref{eq:GMM_all_three}.
Finally, the danger index attributed to a position with respect to the model based on GMR is defined as 
\begin{equation} \label{eq:DI_GMR}
\DI_{\GMR}(q_{ln},q_{lt})=\int_{q_{lt}-\varepsilon}^{q_{lt}+\varepsilon}f_{\GMR}(p_{lt} \mid q_{ln}) dp_{lt},
\end{equation}
in which $f_{\GMR}(p_{lt} \mid p_{ln})$ is given in~\eqref{eq:pdf_GMR_all_three}.

Relations~\eqref{eq:DI_lin_sto_model} and~\eqref{eq:DI_GMR} can be now used to compute the danger index of a path. Let $\pi(x_s,x_e)$ be the path connecting states $x_s$ and $x_e$ and let $\Delta\pi(x_s,x_e)=(x_1,x_2,\dots,x_N)$ 
denote a uniform discretization of the path $\pi(x_s,x_e)$ such that $x^1=x_s$ and $x^N=x_e$. Let $\proj: X \rightarrow Q$ be a function that maps the robot state to the robot configuration. Then, $(q_{ln,i}, q_{lt,i})$ can be defined as the coordinates corresponding to the $i^{th}$ element of $\Delta\pi(x_s,x_e)$, i.e., $(q_{x,i},q_{y,i})=\proj(x_i)$ expressed in the coordinate frame of the human.
The path danger index corresponding to $\pi(x_s,x_e)$ is computed considering linear stochastic model as 
\begin{equation}\label{eq:PDI_lin_sto_model}
\PDI_{SM}(\pi(x_s,x_e))=\sum_{i=1}^{N}\DI_{SM}\left(q_{ln,i}, q_{lt,i}\right),
\end{equation}
in which $\DI_{SM}$ is given in~\eqref{eq:DI_lin_sto_model}, and considering the model based on GMR as
\begin{equation*}
\PDI_{\GMR}(\pi(x_s,x_e))=\sum_{i=1}^{N}\DI_{\GMR}\left(q_{ln,i}, q_{lt,i}\right),
\end{equation*}
in which $\DI_{\GMR}$ is defined in \eqref{eq:DI_GMR}.\\
Suppose the projection of the trajectory $\pi(x_s,x_e)$ on the configuration space results in a line segment, written in the coordinate frame of the pedestrian as 
\begin{equation*}
r(x_s,x_e)=(1-s)\begin{bmatrix}q_{ln,s}\\q_{lt,s}\end{bmatrix}+s\begin{bmatrix}q_{ln,e}\\q_{lt,e}\end{bmatrix}=\begin{bmatrix}r_{ln}(s)\\r_{lt}(s)\end{bmatrix}
\end{equation*}
in which $0<s<1$, and $(q_{ln,s},q_{lt,s})$ and $(q_{ln,e},q_{lt,e})$ are the robot positions corresponding to $x_s$ and $x_e$, respectively, whose coordinates are expressed in the human-fixed reference frame. Then, the \ac{pdf} $f_{\GMM}$, given in~\eqref{eq:GMM_all_three} along the line segment, can be evaluated, which corresponds to the danger index of the path given by
\begin{displaymath}
\PDI_{\GMM}(\pi(x_s,x_e))=
\int_{0}^{1}f_{\GMM}(r_{ln}(s),r_{lt}(s))\left\Vert{\zeta(s)}\right\Vert ds
\end{displaymath}
in which $\left\Vert{\zeta(s)}\right\Vert=\sqrt{\frac{d r_{e_{1}}^{2}(s)}{d s}+\frac{d r_{e_{2}}^{2}(s)}{d s}}$.

\section{Human-aware path planning}
This section describes the proposed approach to human-aware path planning.

\RRTX, a sampling-based algorithm introduced in~\cite{otte2016RRTX}, with asymptotic guarantees on optimality and which allows iterative replanning for time varying obstacle regions, is adopted. %here considered. 
To account for humans, danger indices, computed based on the models described in Section~\ref{sec:hum_models}, are added to the cost function, which attributes a cost to each edge of the tree built by \RRTX. First, \RRTX 
%the adopted path planning algorithm 
is briefly described, and then the proposed cost function is introduced. 

\RRTX~builds a graph $\mathcal{G}=(V,E)$ embedded in $X$, whose nodes $x\in V \subseteq X$ are the states, and directed edges $(v,u) \in E$ correspond to paths that connect a pair of nodes such that $u,v\in V$. The robot starts at $v_{start}$ and goes to $v_{goal}$, if possible. 
$\mathcal{G}$ is built by drawing i.i.d. samples from a uniform distribution over $X$. The algorithm also keeps an \emph{optimal-path subtree} of $\G$, denoted as $\T=(V_\T,E_\T)$ rooted at $v_{goal}$. 
%Furthermore, $v_{goal}=(q_{goal}, 0)$, such that the time component is negative for each node other than the goal. 
%For each node $v$, the algorithm keeps track of incoming and outgoing neighbors corresponding to the edges that $v$ take part in. 
Each edge $(v,u)$ is associated with a positive cost denoted by $\cost_\pi(v,u)$ that captures the path length and the danger of colliding with a human, and is defined as 
\begin{equation}\label{eq:cost_func}
\cost_\pi(u,v)=d_\pi(u,v)\big( 1 + \PDI(\pi(u,v)\big),    
\end{equation}
in which $\PDI(\pi(u,v))$ is the danger index of a path $\pi(u,v)$ computed as described in Section~\ref{sec:danger_index}. 
Each node $v \in V_\T$ is attributed a cost $g(v)$, that is the ($\epsilon$-consistent) cost to reach the goal through $\T$. 
%To each node in $v \in V_\T$ a cost $g(v)$, that is the ($\epsilon$-consistent) cost to reach the goal through $\T$, is associated. 
The estimate of the cost-to-go, that is, $\lmc(v)$, is also stored. It is updated, when appropriate conditions are met, as 
$$
\lmc(v) = \min_{u\in N^+(v)} \cost(v,u) + \lmc(u),
$$
in which $N^+(v)$ is the set of neighbors of $v$ corresponding to outgoing edges. $v$ is $\epsilon-$consistent iff $g(v)-\lmc(v) < \epsilon$. \\
At each iteration, a node $v$ is sampled and is attempted to be connected to $\G$ and $\T$, if possible. If $v$ is connected to the graph, the graph is rewired in the neighborhood of $v$, similar to \RRTst~\cite{karaman2011sampling}. Unlike \RRTst, rewiring one node triggers a cascade of rewiring operations to ensure that the nodes are $\epsilon$-consistent. A priority queue based on the cost-to-go and $\lmc$ is used for rewiring cascades. 

Each human is associated with a \emph{future position region} that captures the area containing the majority of the paths that can be executed by the human within a short time window and denoted by $\mathcal{R}(q_h, \theta_h)$, in which $q_h$ and $\theta_h$ are the position and the orientation of the human, respectively, with respect to a global reference frame. This area is computed using the model given in Section~\ref{sec:hum_model1}, and it corresponds to the convex-hull of the positions whose lateral coordinate is within three standard deviations to either side of the mean. The path danger index of an edge, that is, $\PDI(\pi(u,v))$, is computed only if $\pi(u,v)$ intersects with the future position region. Thus, $\cost(u,v)=d(u,v)$ if $\pi(u,v) \cap \mathcal{R}(q_h, \theta_h) =\emptyset$. In case there are multiple humans within the sensing range, a normalized $\PDI$ is used.
\begin{figure}[htbp]
    \centering
    \subfigure[]{\includegraphics[trim=80 145 100 130,clip,width=0.5\linewidth]{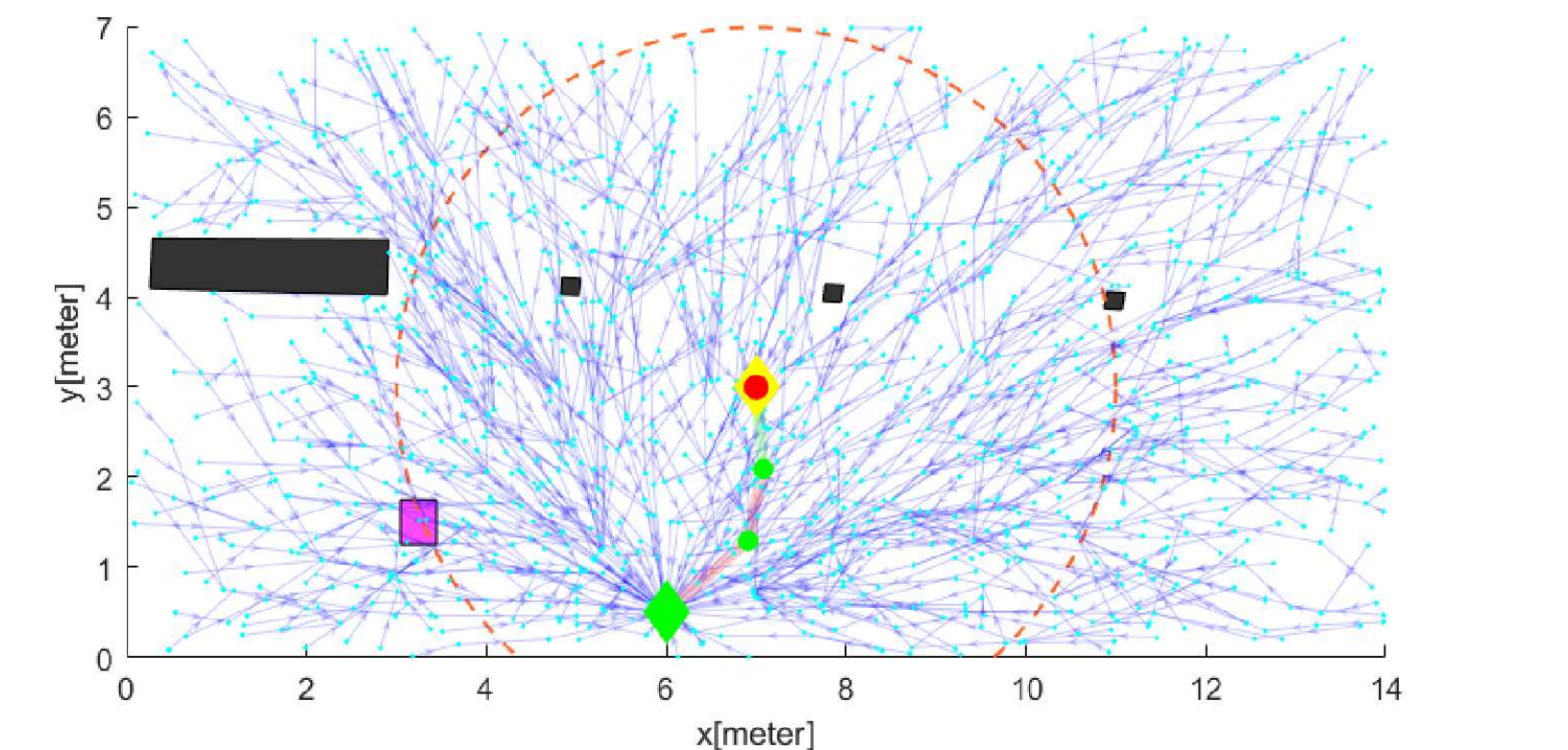}}
    \subfigure[]{\includegraphics[width=0.48\linewidth]{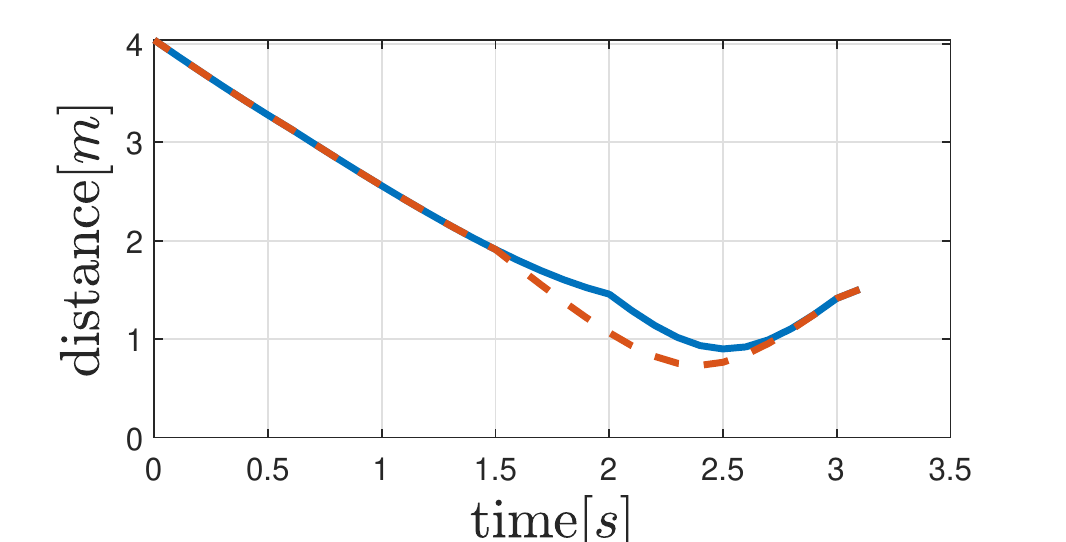}}
    \vspace{-0.75em}
    \caption{One person crossing scenario: (a) initial shortest path, goal is a green diamond, robot is a red dot; (b) human-robot distance, considering danger index as cost (blue line) and minimizing path length only (red line).}
    \label{fig:time_distance}
\end{figure}
\begin{figure*}[thbp]
\centering\hfill
\subfigure[]{\includegraphics[width=.23\linewidth]{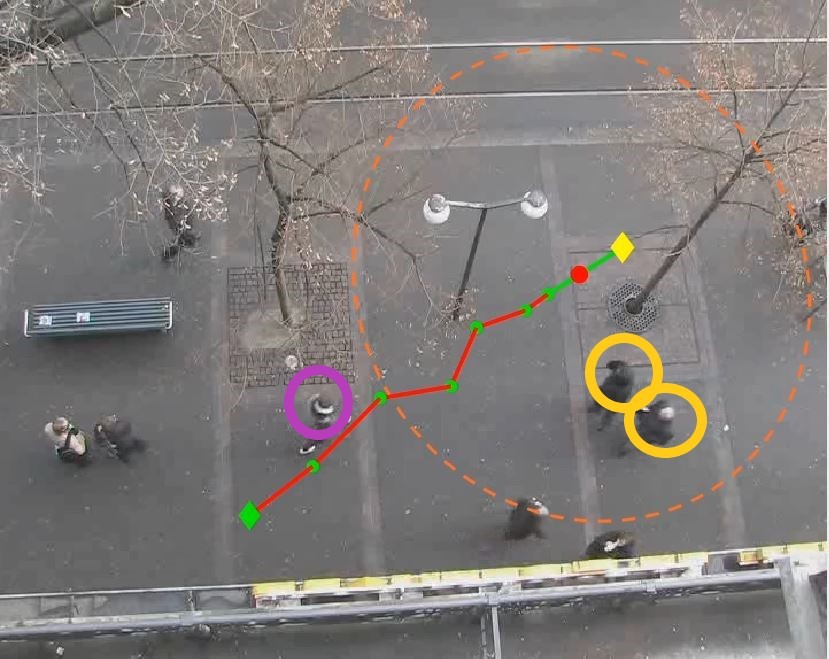}}\hfill
\subfigure[]{\includegraphics[width=.228\linewidth]{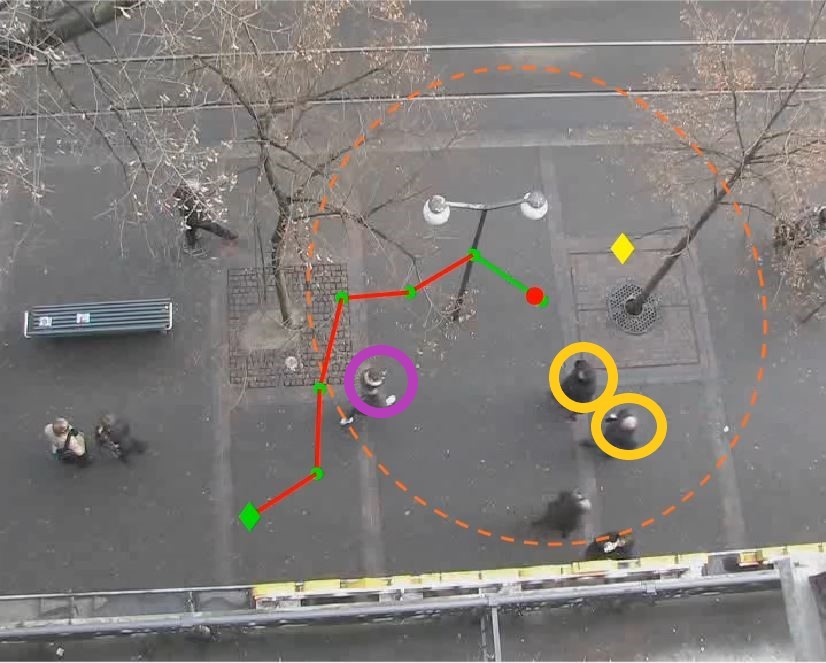}}\hfill 
\subfigure[]{\includegraphics[width=.23\linewidth]{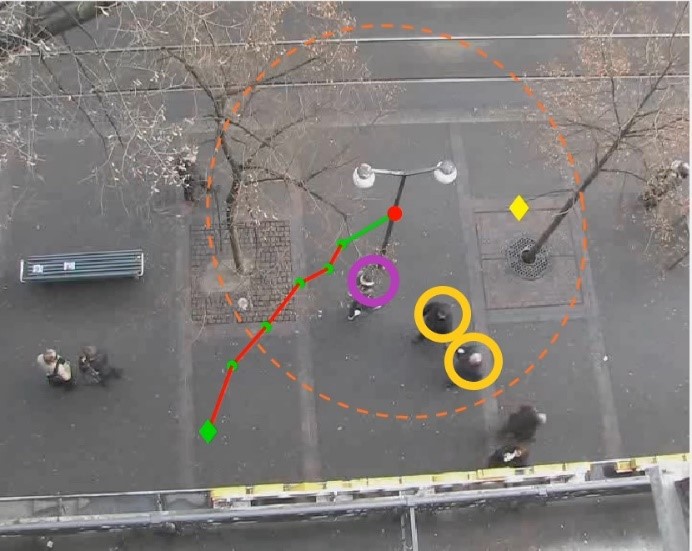}}\hfill\\ \vspace{-0.5em}
\hfill
\subfigure[]{\includegraphics[width=.31\linewidth]{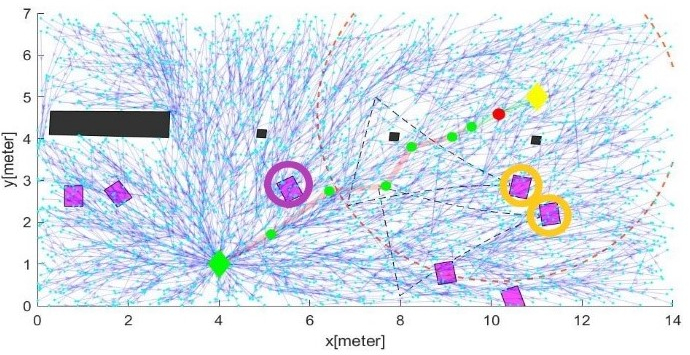}}\hfill
\subfigure[]{\includegraphics[width=.315\linewidth]{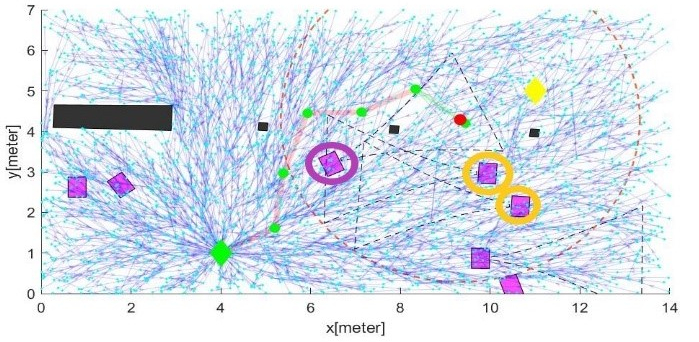}}\hfill
\subfigure[]{\includegraphics[width=.31\linewidth]{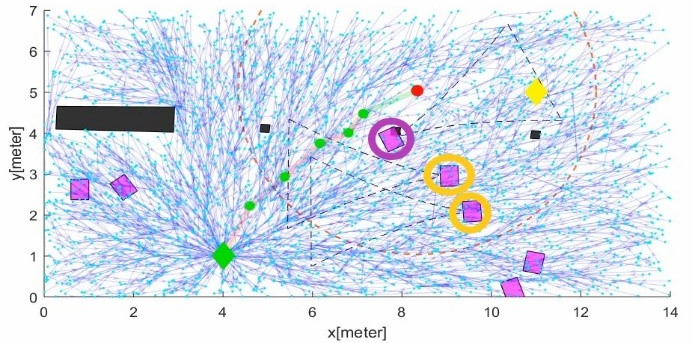}}\hfill
\vspace{-0.75em}
\caption{Instances (from left to right at $0.7$, $1.5$, $2.1$ seconds) from the simulation: (a-c) robot path overlaid on the video frame, (d-f) corresponding graph and minimum-cost trajectory. Robot initial position and goal positions are marked with yellow and green diamonds, respectively.}
\label{fig:sim_results}
\end{figure*}

\section{Simulation results}
This section presents simulation results to show the effectiveness of the proposed approach, and compare the different models used for predicting human motion. 

Simulations are performed on an Intel Core i5-8250U@1.60GHz-1.80GHz CPU personal computer with 8GB RAM. Algorithms are implemented in MATLAB without code optimization. 

The environment used for simulations correspond to a sidewalk in front of a hotel with fixed obstacles, and pedestrians as dynamic obstacles, obtained from BIWI dataset \cite{pellegrini2009you}. This dataset is different than the one used for training. The size of the environment is $7\,\textrm{m}\times14\,\textrm{m}$. Pedestrians are modeled as a rectangle with side lengths equal to $0.5\,\textrm{m}$ and $0.3\,\textrm{m}$, which correspond to the shoulder length and width of an average person, respectively. To avoid inconsistent sampling rates, pedestrian trajectories in the dataset are interpolated using splines and then resampled. 

The robot is equipped with a sensor that can detect pedestrians inside a circular region with radius $4\,\textrm{m}$, that is centered at the current robot position. The robot is assumed to travel at constant speed in between two nodes, and the maximum allowed speed is $2\,\textrm{m/s}$, any edge that violates this condition is considered unfeasible. The robot position and the obstacle set are updated at every $0.1\,\textrm{s}$.

The effectiveness of the approach has been tested using different pairs of initial and goal positions, for all three motion models. No significant differences in the performance of the algorithm employing different prediction models have been observed. This is due to the computed danger indices being very similar.
Two versions of the algorithm, one that employs the cost function defined in \eqref{eq:cost_func} with $\PDI$ defined in \eqref{eq:PDI_lin_sto_model}, and the other one using only the path length, have been also compared. In all the conditions, our method ensured a continuous movement whereas optimizing solely the path-length without considering the potential human motion resulted in stopping the robot to avoid collisions or jittery motion when the path switches between two homotopy classes, in certain scenarios. Furthermore, cost function employing the danger index often results in keeping a larger distance between the robot and the humans. An example is shown in Fig.~\ref{fig:time_distance}a, in which the robot needs to cross a pedestrian moving straight, robot clearance from the pedestrian as a function of time is presented in Fig.~\ref{fig:time_distance}b. Instances from a simulation example corresponding to cost function with $\PDI$ defined in \eqref{eq:PDI_lin_sto_model} is given in Fig.~\ref{fig:sim_results}.

\section{Conclusion}
In this paper a method for safe navigation of a mobile robot in dynamic environments shared with humans is proposed. The method builds upon \RRTX, for which the edge cost is modified to capture the danger of colliding with a human. To estimate this index, two methods are introduced, one based on a linear stochastic model, and another one on a GMM/GMR model. The approach has been validated in simulation, using trajectories recorded from a real-world scenario. Future work includes extending this method to take into account proxemics \cite{edward1966hall} and other social rules.

%\addtolength{\textheight}{-12cm}   % This command serves to balance the column lengths
                                  % on the last page of the document manually. It shortens
                                  % the textheight of the last page by a suitable amount.
                                  % This command does not take effect until the next page
                                  % so it should come on the page before the last. Make
                                  % sure that you do not shorten the textheight too much.

%%%%%%%%%%%%%%%%%%%%%%%%%%%%%%%%%%%%%%%%%%%%%%%%%%%%%%%%%%%%%%%%%%%%%%%%%%%%%%%%

%%%%%%%%%%%%%%%%%%%%%%%%%%%%%%%%%%%%%%%%%%%%%%%%%%%%%%%%%%%%%%%%%%%%%%%%%%%%%%%%

%%%%%%%%%%%%%%%%%%%%%%%%%%%%%%%%%%%%%%%%%%%%%%%%%%%%%%%%%%%%%%%%%%%%%%%%%%%%%%%%
% \section*{APPENDIX}

% Appendixes should appear before the acknowledgment.

\section*{ACKNOWLEDGMENT}
This work was supported by Academy of Finland (project CHiMP 342556).
The authors thank former M.Sc. students Alessandro Botti and H\"{u}seyin Hanl{\i} for their valuable input and for implementing the ideas presented in this paper. 

%%%%%%%%%%%%%%%%%%%%%%%%%%%%%%%%%%%%%%%%%%%%%%%%%%%%%%%%%%%%%%%%%%%%%%%%%%%%%%%%

\bibliographystyle{IEEEtran}
\bibliography{root}

% Generated by IEEEtran.bst, version: 1.14 (2015/08/26)
\begin{thebibliography}{10}
\providecommand{\url}[1]{#1}
\csname url@samestyle\endcsname
\providecommand{\newblock}{\relax}
\providecommand{\bibinfo}[2]{#2}
\providecommand{\BIBentrySTDinterwordspacing}{\spaceskip=0pt\relax}
\providecommand{\BIBentryALTinterwordstretchfactor}{4}
\providecommand{\BIBentryALTinterwordspacing}{\spaceskip=\fontdimen2\font plus
\BIBentryALTinterwordstretchfactor\fontdimen3\font minus
  \fontdimen4\font\relax}
\providecommand{\BIBforeignlanguage}[2]{{%
\expandafter\ifx\csname l@#1\endcsname\relax
\typeout{** WARNING: IEEEtran.bst: No hyphenation pattern has been}%
\typeout{** loaded for the language `#1'. Using the pattern for}%
\typeout{** the default language instead.}%
\else
\language=\csname l@#1\endcsname
\fi
#2}}
\providecommand{\BIBdecl}{\relax}
\BIBdecl

\bibitem{otte2016RRTX}
M.~Otte and E.~Frazzoli, ``{RRTX}: Asymptotically optimal single-query
  sampling-based motion planning with quick replanning,'' \emph{Int. J. Rob.
  Res.}, vol.~35, no.~7, pp. 797--822, 2016.

\bibitem{de2007onboard}
G.~De~Nicolao, A.~Ferrara, and L.~Giacomini, ``Onboard sensor-based collision
  risk assessment to improve pedestrians' safety,'' \emph{IEEE Tran. Vehic.
  Tech.}, vol.~56, no.~5, pp. 2405--2413, 2007.

\bibitem{calinon2007learning}
S.~Calinon, F.~Guenter, and A.~Billard, ``On learning, representing, and
  generalizing a task in a humanoid robot,'' \emph{IEEE Tran. Sys., Man, Cyb.,
  Part B}, vol.~37, no.~2, pp. 286--298, 2007.

\bibitem{stentz1995focussed}
A.~Stentz \emph{et~al.}, ``The focussed d\^{}* algorithm for real-time
  replanning,'' in \emph{IJCAI}, vol.~95, 1995, pp. 1652--1659.

\bibitem{koenig2004lifelong}
S.~Koenig, M.~Likhachev, and D.~Furcy, ``Lifelong planning a\^{}*,''
  \emph{Artificial Intelligence}, vol. 155, no. 1-2, pp. 93--146, 2004.

\bibitem{koenig2005fast}
S.~Koenig and M.~Likhachev, ``Fast replanning for navigation in unknown
  terrain,'' \emph{IEEE Tran. Rob.}, vol.~21, no.~3, pp. 354--363, 2005.

\bibitem{ferguson2006replanning}
D.~Ferguson, N.~Kalra, and A.~Stentz, ``Replanning with {RRTs},'' in \emph{Int.
  Conf. on Rob. Autom.}, 2006, pp. 1243--1248.

\bibitem{zucker2007multipartite}
M.~Zucker, J.~Kuffner, and M.~Branicky, ``Multipartite {RRTs} for rapid
  replanning in dynamic environments,'' in \emph{Int. Conf. on Rob. Autom.},
  2007, pp. 1603--1609.

\bibitem{adiyatov2017novel}
O.~Adiyatov and H.~A. Varol, ``A novel {RRT*}-based algorithm for motion
  planning in dynamic environments,'' in \emph{Int. Conf. on Mechatronics and
  Automation}, 2017, pp. 1416--1421.

\bibitem{connell2017dynamic}
D.~Connell and H.~M. La, ``Dynamic path planning and replanning for mobile
  robots using {RRT},'' in \emph{Int. Conf. on Systems, Man, and Cybernetics},
  2017, pp. 1429--1434.

\bibitem{geraerts2010planning}
R.~Geraerts, ``Planning short paths with clearance using explicit corridors,''
  in \emph{Int. Conf. on Rob. Autom.}, 2010, pp. 1997--2004.

\bibitem{wang2020neural}
J.~Wang, W.~Chi, C.~Li, C.~Wang, and M.~Q.-H. Meng, ``Neural {RRT*}:
  Learning-based optimal path planning,'' \emph{IEEE Tran. Autom. Sc. Eng.},
  vol.~17, no.~4, pp. 1748--1758, 2020.

\bibitem{denny2014marrt}
J.~Denny, E.~Greco, S.~Thomas, and N.~M. Amato, ``{MARRT}: Medial axis biased
  rapidly-exploring random trees,'' in \emph{Int. Conf. on Rob. Autom.}, 2014,
  pp. 90--97.

\bibitem{sakcak2021complete}
B.~Sakcak and S.~M. LaValle, ``Complete path planning that simultaneously
  optimizes length and clearance,'' in \emph{Int. Conf. on Rob. Autom.}, 2021,
  pp. 10\,100--10\,106.

\bibitem{van2008planning}
J.~Van Den~Berg and M.~Overmars, ``Planning the shortest safe path amidst
  unpredictably moving obstacles,'' in \emph{Algorithmic Foundation of Robotics
  VII}.\hskip 1em plus 0.5em minus 0.4em\relax Springer, 2008, pp. 103--118.

\bibitem{luders2010chance}
B.~Luders, M.~Kothari, and J.~How, ``Chance constrained {RRT} for probabilistic
  robustness to environmental uncertainty,'' in \emph{AIAA guidance,
  navigation, and control conference}, 2010, p. 8160.

\bibitem{rudenko2020human}
A.~Rudenko, L.~Palmieri, M.~Herman, K.~M. Kitani, D.~M. Gavrila, and K.~O.
  Arras, ``Human motion trajectory prediction: A survey,'' \emph{Int. J. Rob.
  Res.}, vol.~39, no.~8, pp. 895--935, 2020.

\bibitem{pellegrini2009you}
S.~Pellegrini, A.~Ess, K.~Schindler, and L.~Van~Gool, ``You'll never walk
  alone: Modeling social behavior for multi-target tracking,'' in \emph{Int.
  Conf. on Computer Vision}, 2009, pp. 261--268.

\bibitem{karaman2011sampling}
S.~Karaman and E.~Frazzoli, ``Sampling-based algorithms for optimal motion
  planning,'' \emph{Int. J. Rob. Res.}, vol.~30, no.~7, pp. 846--894, 2011.

\bibitem{edward1966hall}
E.~T. Hall, \emph{The Hidden Dimension}.\hskip 1em plus 0.5em minus 0.4em\relax
  Doubleday, 1966.

\end{thebibliography}

\end{document}